\pdfoutput=1

\documentclass[11pt]{article}

\usepackage{acl}

\usepackage{times}
\usepackage{latexsym}

\usepackage{todonotes}
\usepackage{amsmath} 
\usepackage{enumitem}
\usepackage{comment}
\usepackage{multirow}
\usepackage{booktabs}
\usepackage{float}
\usepackage{xcolor}
\usepackage{array}
\usepackage{tabularx}

\definecolor{verde}{rgb}{0.0, 0.5, 0.0}

\newcolumntype{C}[1]{>{\centering\arraybackslash}p{#1}}

\usepackage[T1]{fontenc}

\usepackage[utf8]{inputenc}

\usepackage{microtype}

\usepackage{inconsolata}

\title{Investigating the Frequency Distortion of Word Embeddings and Its Impact on Bias Metrics}

\author{
    Francisco Valentini$^{1,2}$ \,
    Juan Cruz Sosa$^{3}$ \,
    Diego Fernandez Slezak$^{1,3}$ 
    Edgar Altszyler$^{1,2}$ \\ \\[-.1cm]
    $^{1}$Instituto de Investigación en Ciencias de la Computación, CONICET-UBA, Argentina \\
    $^{2}$Maestría en Data Mining, Universidad de Buenos Aires (UBA), Argentina \\
    $^{3}$Departamento de Computación, FCEyN, UBA, Argentina \\
    \small{\texttt{fvalentini@dc.uba.ar}},\, 
    \small{\texttt{juan.cruz.sosa.92@gmail.com}},\, 
    \small{\texttt{dfslezak@dc.uba.ar}},\, 
    \small{\texttt{ealtszyler@dc.uba.ar}}\\
}

\begin{document}
\maketitle
\begin{abstract}
Recent research has shown that static word embeddings can encode words' frequencies. However, little has been studied about this behavior. In the present work, we study how frequency and semantic similarity relate to one another in static word embeddings, and we assess the impact of this relationship on embedding-based bias metrics. We find that Skip-gram, GloVe and FastText embeddings tend to produce higher similarity between high-frequency words than between other frequency combinations. We show that the association between frequency and similarity also appears when words are randomly shuffled, and holds for different hyperparameter settings. This proves that the patterns we find are neither due to real semantic associations nor to specific parameters choices, and are an artifact produced by the word embeddings. To illustrate how frequencies can affect the measurement of biases related to gender, ethnicity, and affluence, we carry out a controlled experiment that shows that biases can even change sign or reverse their order when word frequencies change.\footnote{Code for the paper is available at \url{https://github.com/ftvalentini/EmbeddingsFrequency}}
\end{abstract}

\section{Introduction} \label{sec:introduction}

Static word embeddings have proven to encode semantic information of words and are therefore useful to solve tasks such as synonym selection and analogical reasoning \citep{mikolov2013distributed, levy2015improving}. More recent contextualized representations have achieved better results \citep{ethayarajh2019contextual}, specially in tasks where the local context of words is important \citep{sezerer2021survey}. However, static word embeddings are still widely used in computational social science studies that examine global aspects of corpora. For example, embeddings are trained on specific corpora and are used to compute metrics that quantify societal biases and stereotypes that might be present in the text \citep{garg2018word, kozlowski2019geometry, defranza2020language, jones2020stereotypical, lewis2020gender, charlesworth2021gender}. Static embeddings are also used in a wide range of applications like topic coherence evaluation \cite{aletras2013evaluating}, dream theory analysis \cite{altszyler2017interpretation}, literature studies \cite{diuk2012quantitative}, and cognitive science studies \cite{mota2022imagetic}. 

Previous research has found static word embeddings appear to be associated with word frequency in various ways: word frequency correlates with embedding norm \citep{wilson2015controlled, arora2016latent}, the nearest neighbors of the embeddings of medium-frequency English words are more unstable \citep{hellrich2016bad}, there are frequency-related differences in the distribution of the inner products between target and context vectors \citep{mimno2017strange}, embeddings can accurately predict whether a word is frequent or rare \citep{schnabel2015evaluation}, and the visual inspection of their top principal components suggest they encode frequency \citep{gong2018frage, mu2018allbutthetop}. When it comes to using embeddings to measure bias in text, \citet{valentini2022undesirable} found that gender embedding-based bias metrics can spuriously depend on word frequency. 

Our work addresses several gaps in the existing literature regarding the frequency distortion of static word embeddings and its impact on the quantification of biases in corpora. Even if it has been pointed out that embeddings can encode frequency, this is the first study that:

\begin{enumerate}[leftmargin=*,noitemsep,topsep=0pt]
    \item Comprehensively investigates the association between frequency and similarity in commonly used embeddings.
    \item Examines whether embeddings encode frequency due to undesirable properties of embeddings or actual properties of corpora.
    \item Explores the persistence of the frequency distortion under different hyperparameter settings.
    \item Assesses the impact on a computational social science application, namely bias measurement. 
\end{enumerate}

\begin{figure*}[!t]
    \centering
    \includegraphics[width=\linewidth]{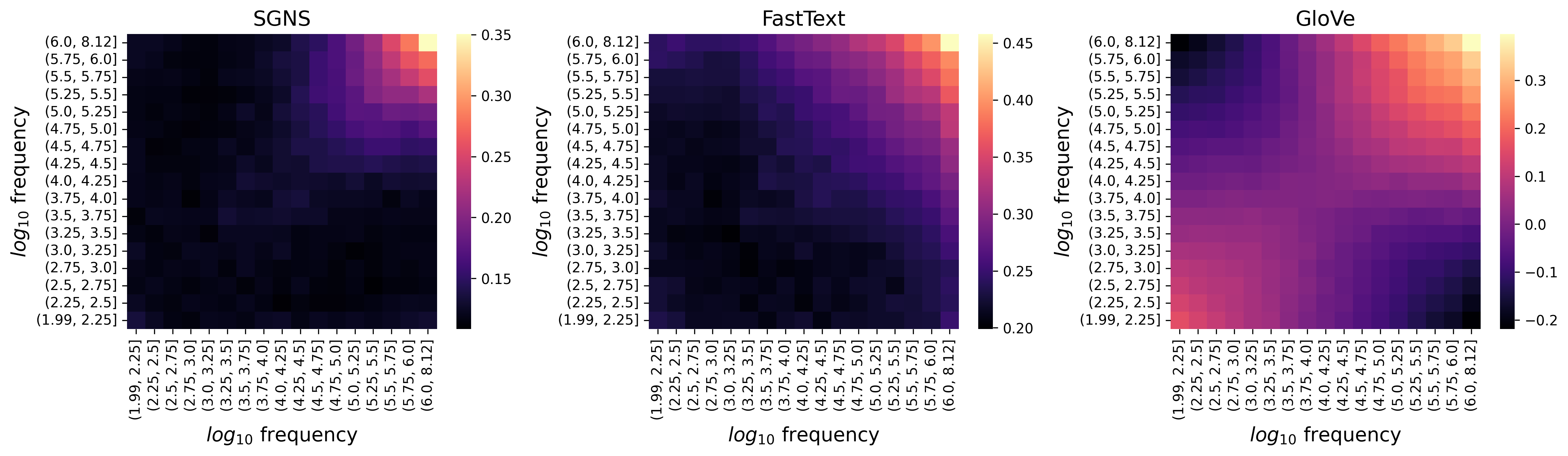}
    \caption{Mean cosine similarity between 500 random word pairs for each combination of frequencies, in embeddings trained on Wikipedia. \textbf{Cosine similarity tends to be higher between high frequency words.}}
    \label{fig:cosine_unshuffled}
\end{figure*}

\begin{figure*}[!t]
    \centering
    \includegraphics[width=\linewidth]{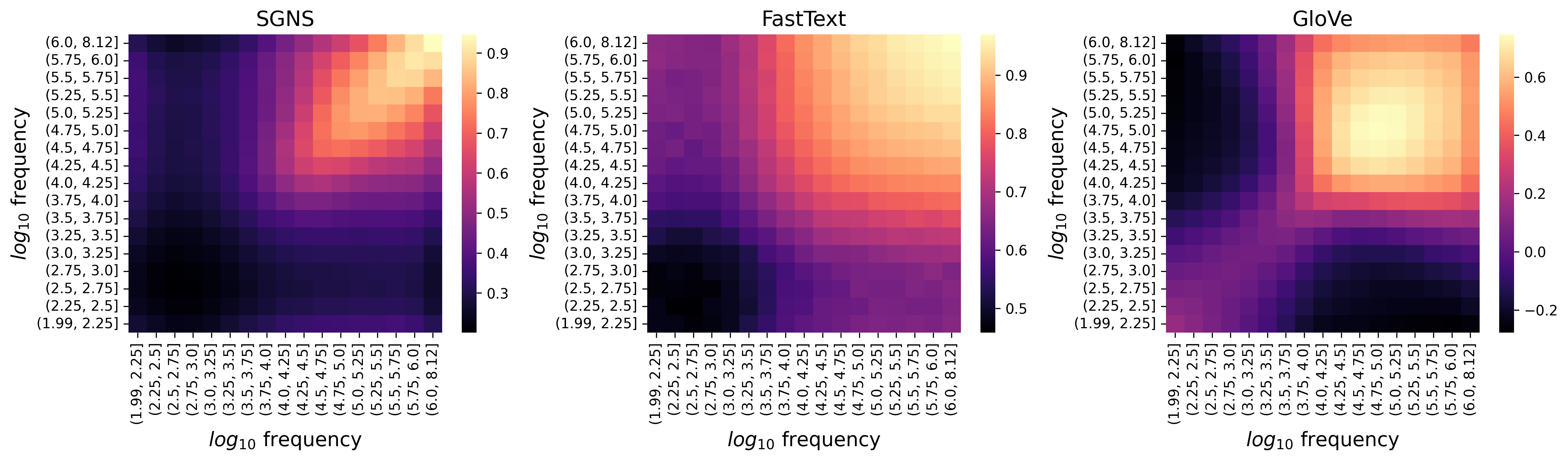}
    \caption{Mean cosine similarity between 500 random word pairs for each combination of frequencies, in embeddings trained on a shuffled version of Wikipedia. \textbf{The similarity of embeddings trained on a corpus with random co-occurrences depends on the words' frequencies.}}
    \label{fig:cosine_shuffled}
\end{figure*}


\section{The dependence of word embeddings on frequency} \label{sec:dependence}

As a first step, we seek to describe the association between word frequency and cosine similarity, commonly used to measure semantic closeness.

We use the 2021 English Wikipedia as a corpus to train embeddings with word2vec with skip-gram with negative-sampling (SGNS, \citealp{mikolov2013distributed}), GloVe \citep{pennington2014glove} and FastText \citep{bojanowski2017enriching} with default hyperparameters (see details in Appendix \ref{app:methods}). We group all words by their $\log_{10}$ frequency, and use matrices to represent the mean cosine similarity of 500 randomly sampled pairs of words for each combination of bins, excluding comparisons between the same word.  

In the three methods the mean cosine similarity is higher between high frequency words than between any other combination of frequencies (Figure \ref{fig:cosine_unshuffled}). Unlike SGNS and FastText, the GloVe mean similarity is moderate to high between words of the same frequency range (the matrix diagonal) and low between words of different frequencies (e.g. words with $10^6$ vs $10^2$ frequencies). 

These results seem to imply that word frequencies influence how similar two words are, raising the following questions: is this due to an artifact of the embeddings? Or does it reveal actual properties of the corpus; for example, that high-frequency words are actually semantically closer on average to one another than the rest of the vocabulary? We conduct the following study to answer this.

\subsection{Experiments} \label{sec:experiments}

Following \citet{valentini2022undesirable}'s approach, we produce a randomly shuffled Wikipedia corpus, with tokens distributed at random across the text. As co-occurrences are random, words retain their frequency but any contextual information is lost. We train embeddings on this corpus and repeat the analysis from the previous section (as in Figure \ref{fig:cosine_unshuffled}). 

If any association is found between similarity and frequency in this setting, it should be explained only by word frequencies. If embeddings don't capture frequency, we would expect a uniform distribution of cosine similarity across all frequency combinations: the similarity of any pair of words should be on average the same.

We find that the mean cosine similarities of embeddings trained on the shuffled corpus depend on the frequencies of the words, and this happens in different ways depending on the method (Figure \ref{fig:cosine_shuffled}). When comparing frequent words (frequency around $10^4$ and above) to one another, all embeddings tend to yield high similarities; and similarity tends to drop in different ways when doing other comparisons. 

We obtain the same qualitative result when using an Euclidean distance-based similarity measure: the similarity of any two words depends heavily on their frequencies (Figures \ref{fig:euclidian_unshuffled} and \ref{fig:euclidian_shuffled} in Appendix \ref{app:euclidean}). Therefore the frequency-based effect is not caused by the choice of cosine similarity. 

\begin{figure}
    \centering
    \includegraphics[width=\linewidth]{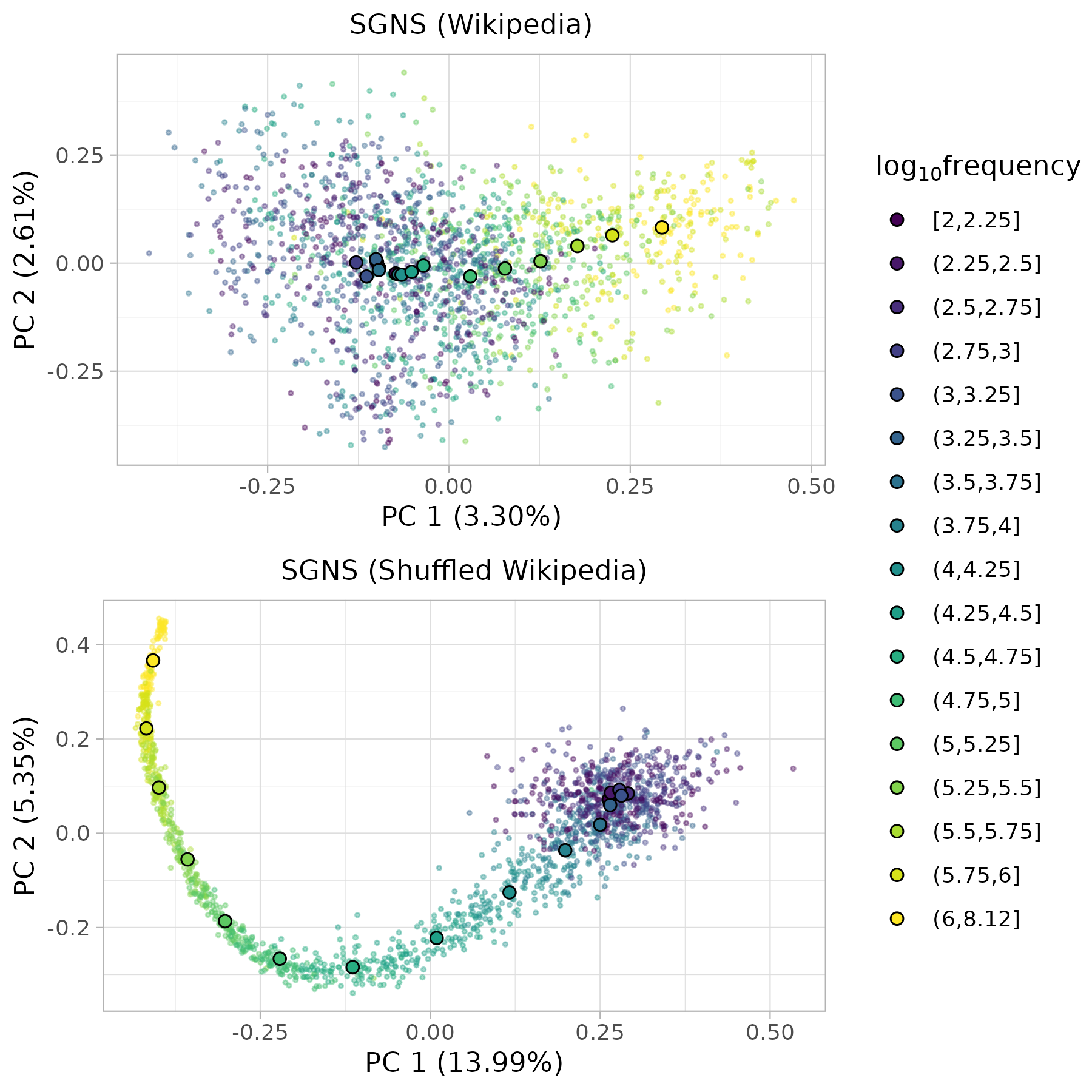}
    \caption{
        PCA of a sample of SGNS embeddings stratified by frequency, trained on Wikipedia (top) and shuffled Wikipedia (bottom). Vectors are normalized to unit length before PCA and there are 100 words by frequency bin. Centroids are displayed with larger markers. Figure \ref{fig:pca_centroids} in Appendix \ref{app:pca} displays the plots for GloVe and FastText. \textbf{The top two components are highly associated with the frequency dimension: the geometry of vectors encodes the training corpus frequencies.}
    }
    \label{fig:pca_sgns}
\end{figure}

The fact that shuffling words prior to training does not yield a uniform distribution of similarity across frequencies suggests that embeddings tend to encode frequency. To further assess this we do PCA on the vectors of a sample of words stratified by frequency and inspect the centroids of the top two components of each frequency bin (top panel of Figure \ref{fig:pca_sgns}). PCA finds the dimensions with the most variability, and we find that the top two are highly associated with the frequency dimension. Therefore the geometry of vectors trained on the original corpus encodes training data frequencies, which is consistent with the literature's previous findings.

We also run PCA on the vectors trained on the shuffled corpus to confirm that this is not a result of properties of the corpus but rather an artifact of embeddings. The trend is more pronounced in this setting: words with varying frequency tend to live in distinct regions in the embedding space (bottom panel of Figure \ref{fig:pca_sgns}). Thus embedding-based similarity metrics can detect semantic closeness even when there shouldn't be any. Four additional independent random shuffles of the corpus yielded the same qualitative results.

PCA in the shuffled corpus also reveals that the vectors of low frequency words are more spread out and therefore have lower similarity between themselves (as seen in the SGNS heatmap of Figure \ref{fig:cosine_shuffled}). This might occur because the distribution of co-occurrences of low-frequency words is less varied, and thus noisier, even when shuffling the corpus. On the other hand, the co-occurrences of higher frequency words are more broadly distributed, and as they are completely random because of the shuffling, the vectors of these words are affected almost exclusively by the frequency dimension, and thus the high similarity between them. 

\begin{figure}
    \centering
    \includegraphics[width=\linewidth]{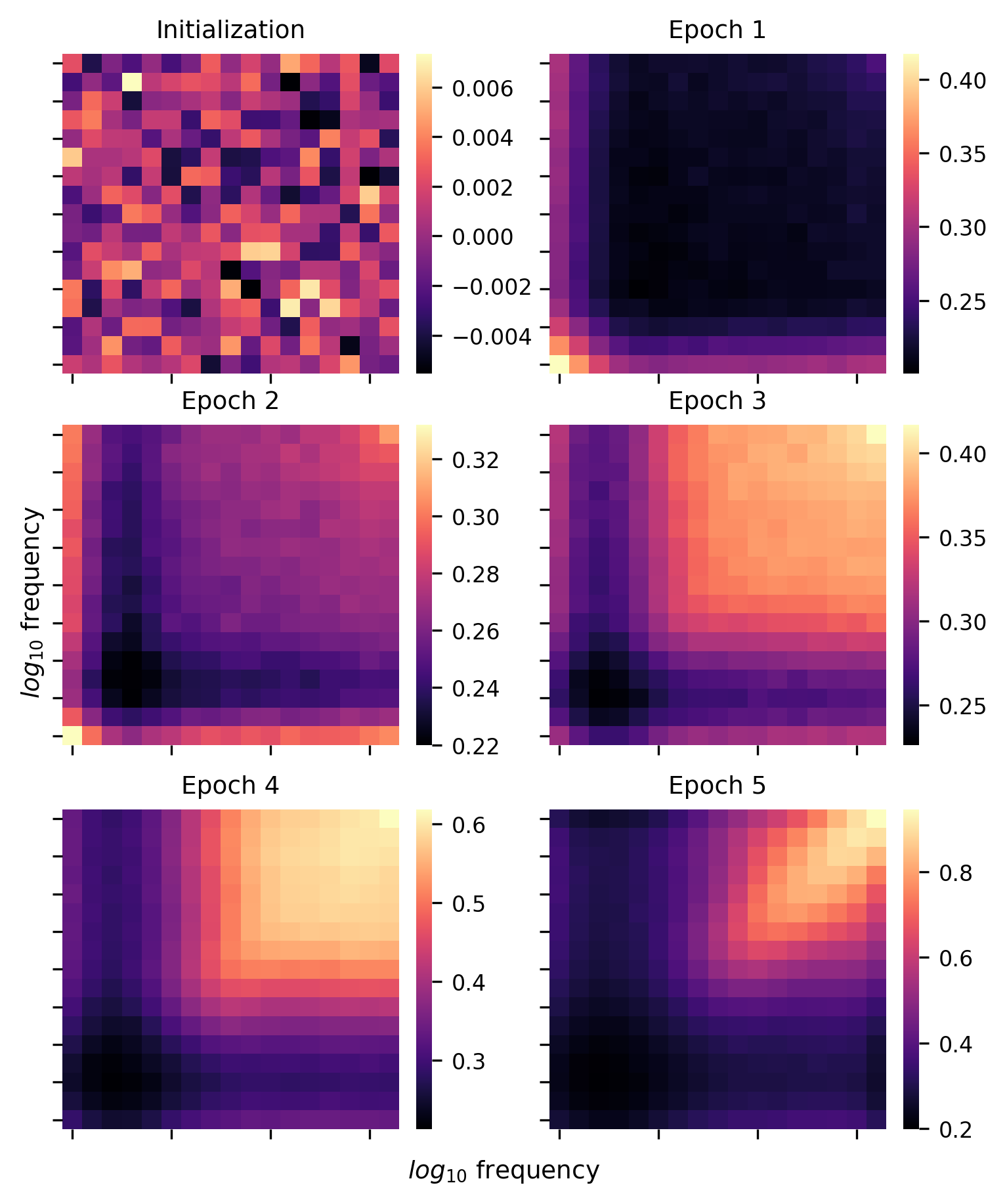}
    \caption{Mean cosine similarity by epoch between 500 random word pairs in SGNS embeddings trained on shuffled Wikipedia. Frequency bins are the same as in Figure \ref{fig:cosine_shuffled}. \textbf{As training progresses, the frequency dimension becomes more salient and the importance of co-occurrences decreases.} }
    \label{fig:sgns_epochs}
\end{figure}
    
We provide an intuition on how this phenomenon arises during training by visualizing the similarity heatmaps of SGNS embeddings of the shuffled corpus in each epoch (Figure \ref{fig:sgns_epochs}). By the end of the first epoch, the vectors of low frequency words have probably moved very little and remain close to their initialization--thus the relatively high similarity between them. In contrast, the vectors of more frequent words are updated with higher variability as their random co-occurrences with other words are more broadly distributed. As training progresses, the frequency dimension becomes more salient and the importance of co-occurrences decreases, so much so that by the last epoch frequency is the dimension that captures the most variance in the vectors (PCA in Figure \ref{fig:pca_sgns}).

\subsection{Sensitivity to hyperparameters}

We evaluate the robustness of the findings from section \ref{sec:experiments} to embeddings' hyperparameters choices. Following \citet{levy2015improving}, the hyperparameters values we explore in all methods are: 
\begin{itemize}[noitemsep]
   \item Window size (\texttt{win}): 2, 5, 10
   \item Adding context vectors (\texttt{w+c}): \emph{yes}, \emph{no}
\end{itemize}
For SGNS and FastText we also explore the following hyperparameters:
\begin{itemize}[noitemsep]
   \item Context distribution smoothing (\texttt{cds}): 0.75, 1
   \item Number of negative samples (\texttt{neg}): 1, 5, 15
\end{itemize}

Trying all combinations of these hyperparameters results in 6 GloVe, 36 SGNS and 36 FastText settings. With each setting we train embeddings on the shuffled corpus.

In order to measure the association between word frequencies and similarity between embeddings, we compute the root mean squared error (RMSE) between the values of each cell of the similarity heatmap and the overall average. The overall average represents the mean similarity we would see in each cell of the heatmap if there were no association between frequencies and similarity. The larger the RMSE, the larger the deviation from the uniform distribution of similarity across frequency combinations, and thus the stronger the association between frequencies and similarity.

As a control, we compare the actual distribution of RMSE with the distribution of RMSE if there was no frequency effect. We compute this by randomly shuffling the similarity values of the pairs of words used to build the heatmap in each hyperparameter setting. The RMSEs would be close to zero in this case, as the random shuffling would result in a uniform distribution of similarity values across frequencies i.e. no association between frequencies and similarity. 


\begin{figure}
    \centering
    \includegraphics[width=\linewidth]{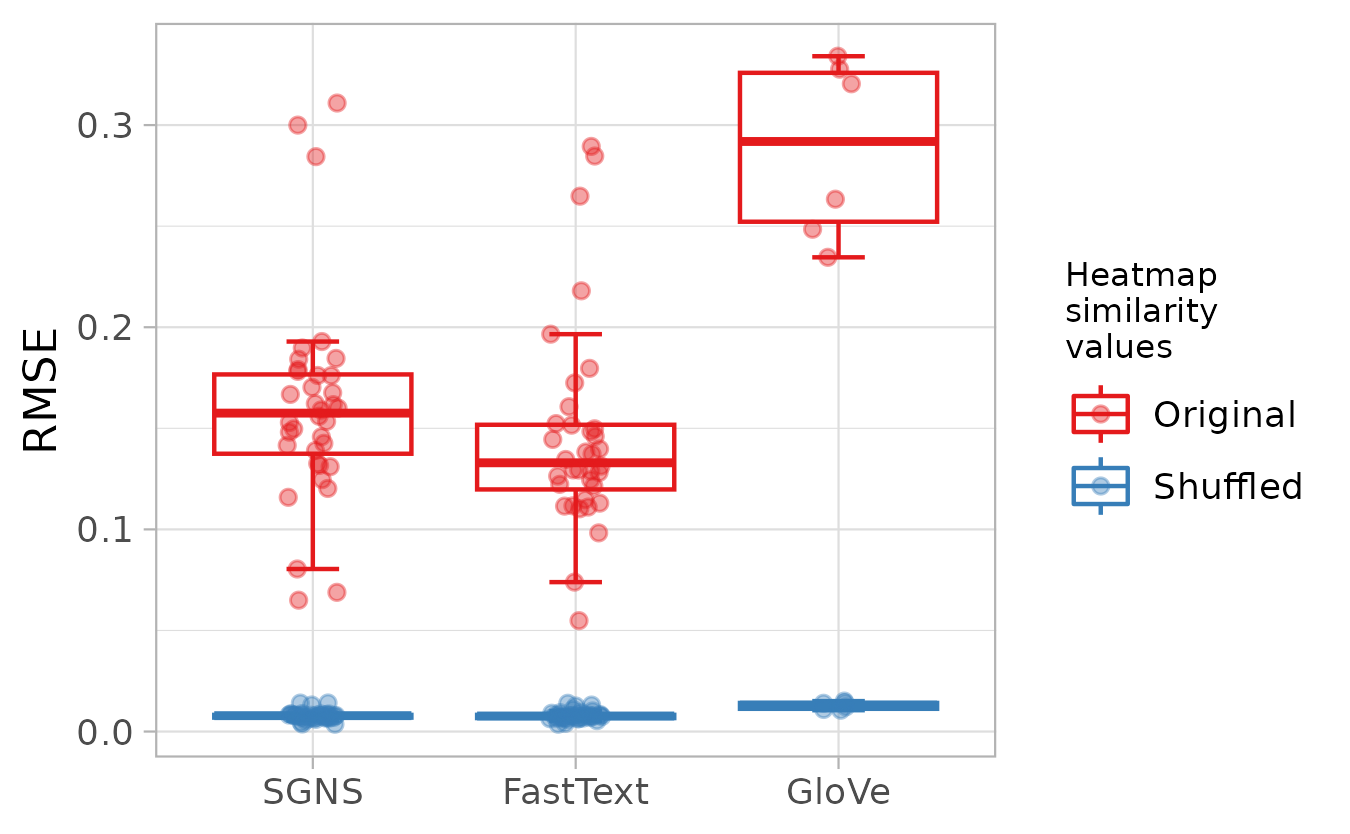}
    \caption{
    Association between frequency and cosine similarity as measured by the RMSE metric. Each point is a hyperparameter setting of embeddings trained on shuffled Wikipedia. Comparing the RMSE computed with the actual heatmaps (red) with the RMSE computed with heatmaps obtained by shuffling the similarity values of the pairs of words (blue) shows that the metric yields low values when there is no relationship between similarity and frequency. \textbf{Word similarity depends on word frequency in all hyperparameter settings.}
    }
    \label{fig:hyperparams_cosine}
\end{figure}

We compute the RMSE metric for cosine similarity in each hyperparameter setting (Figure \ref{fig:hyperparams_cosine}). The RMSE metric is high in all the hyperparameter settings, as compared to the distribution of RMSE obtained by shuffling the similarity values of the pairs of words. This proves that the finding that word similarity depends on word frequencies is robust to hyperparameter choices. In Appendix \ref{app:euclidean} we show this is also the case for an Euclidean distance-based similarity measure.

A linear regression analysis that includes the hyperparameter values as predictors and the RMSE as the outcome variable shows that no specific hyperparameter tends to systematically yield higher or lower RMSE values when using cosine similarity. Only adding context vectors (\texttt{w+c} = \emph{yes}) is significantly associated with a greater association between frequency and Euclidean distance (p-value < 0.0001) in all three methods. Refer to Appendix \ref{app:hyperparameters} for more details.

\section{Assessing the impact} \label{sec:impact}

In computational social science, static word embeddings are typically used to measure societal biases and stereotypes potentially present in corpora. Here we study the effect of the dependence of embeddings on frequency on this type of studies. We highlight that this is different from measuring or mitigating biases in NLP models. Our goal is to assess how much the individual frequency of words might distort the estimates of biases in specific corpora when using word embeddings.

To measure the bias of a target word $x$ we use the difference between the mean similarity of words of context groups $A$ and $B$ with respect to $x$:
\begin{equation} \label{eq:BiasWE} 
    \text{Bias}_{\text{WE}} = 
        \underset{a \in A}{\mathrm{mean}} \; \text{cos}(w_x,w_a) - 
        \underset{b \in B}{\mathrm{mean}} \; \text{cos}(w_x,w_b)
\end{equation}
where $w_i$ is the embedding of word $i$ and $\text{cos}(w_i,w_j)$ is the cosine similarity. $A$ and $B$ are set based on the bias to be measured. For instance, to quantify binary gender bias (female/male), gendered nouns and pronouns are used. 

Here we use the same bias metric as in \citet{lewis2020gender} and \citet{valentini2022undesirable} for its simplicity. Other similar metrics have been used \citep{bolukbasi2016man,garg2018word,kozlowski2019geometry,jones2020stereotypical} and have shown to yield similar results \citep{garg2018word}.


\begin{figure*}[!ht]
    \centering
    \includegraphics[width=\linewidth]{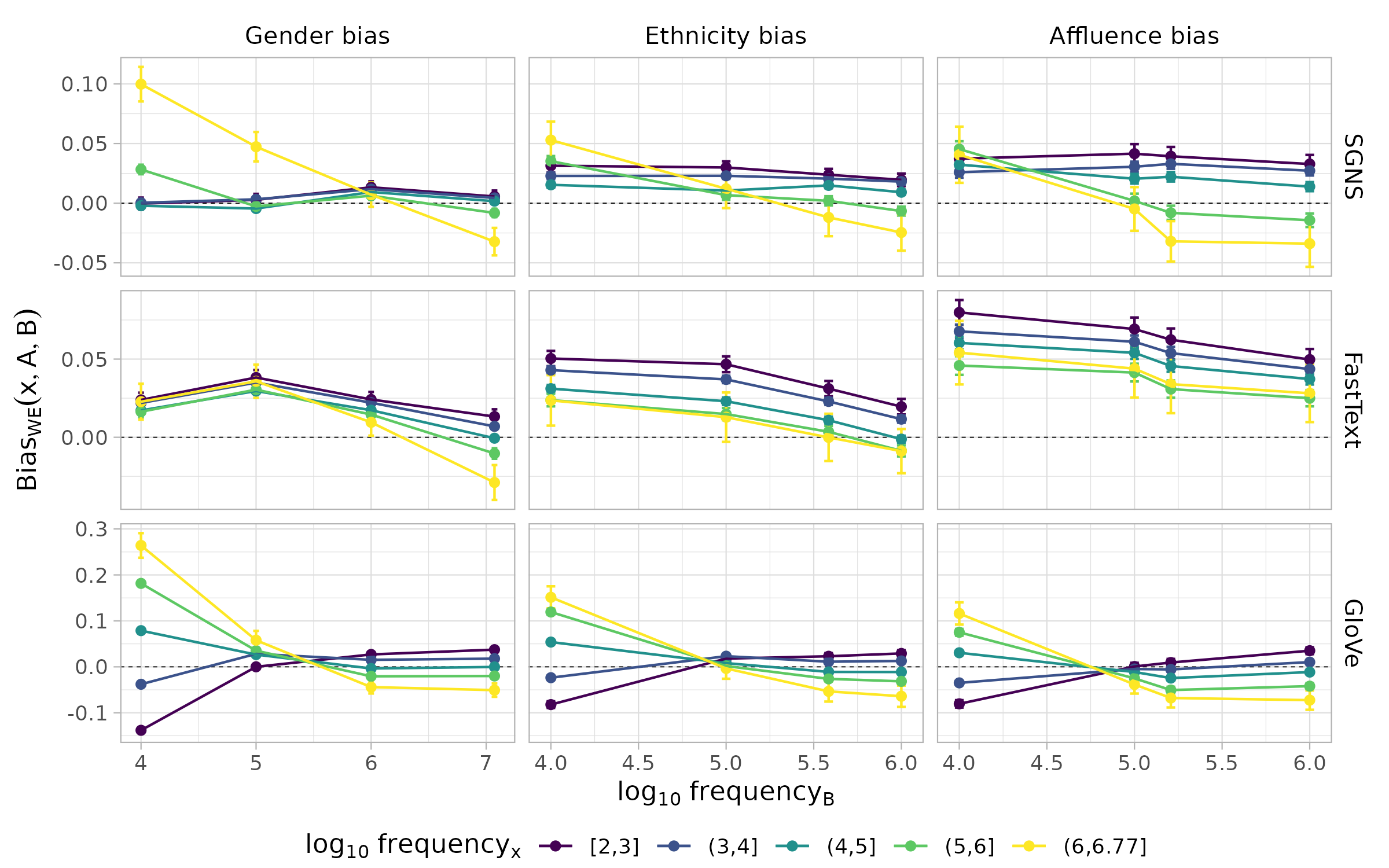}
    \caption{
        Mean biases of selected words grouped by frequency. Horizontal axes represent corpora where the frequency of $B$ varies and the frequency of $A$ is almost constant (see Tables \ref{tab:undersampling_freqs_he}, \ref{tab:resampling_freqs_race} and \ref{tab:resampling_freqs_affluence}). Averages are plotted with bootstrap confidence intervals. Gender, ethnicity, and affluence biases are measured in 4,384, 4,641, and 4,617 target words, respectively.
        \textbf{Changes in word frequencies can generate substantial changes in estimates of gender, ethnicity and affluence bias, even if the underlying distribution of co-occurrences remains constant.}
    }
    \label{fig:impact_bias}
\end{figure*}


\subsection{Experimental setup} \label{sec:impact_setup}

To measure the sensitivity of embedding-based bias to changes in the context words frequencies, we first use the female/male gender bias as a widely studied test case \citep{garg2018word, kozlowski2019geometry, defranza2020language, jones2020stereotypical, lewis2020gender, charlesworth2021gender}. 

We measure gender bias with equation \ref{eq:BiasWE} with $A=\{\textit{she}\}$ and $B=\{\textit{he}\}$ \citep{bolukbasi2016man}. We seek to train embeddings on corpora where one of the words ($A$) has the original corpus' frequency, while the other word ($B$) has a target frequency level. To achieve this, we randomly drop sentences containing $B$ until the target frequency is reached. We set three target frequencies for $B$: $10^4$, $10^5$, and $10^6$. This implies creating three resampled corpora (see Table \ref{tab:undersampling_freqs_he}).

\begin{table}[H]
    \centering
    \begin{tabular}{|p{2.9cm}|C{1.7cm}|C{1.7cm}|}
        \hline
                                      & {\small \emph{she} ($A$)} & {\small \emph{he}} ($B$) \\ \hline
        {\small Original Wiki.  }     & $10^{6.55}$               & $10^{7.07}$              \\ \hline
        {\small Undersampled Wiki. 1} & $10^{6.53}$               & $10^{6}$                 \\ \hline
        {\small Undersampled Wiki. 2} & $10^{6.52}$               & $10^{5}$                 \\ \hline
        {\small Undersampled Wiki. 3} & $10^{6.52}$               & $10^{4}$                 \\ \hline
    \end{tabular}
    \caption{
        Frequency of context words in the undersampling experiment that drops sentences with word $B$ (\emph{he}). The frequency of $A$ (\emph{she}) decreases but to a minor extent.
        }
    \label{tab:undersampling_freqs_he}
\end{table}

Using only one word in each context group to measure bias (as in \citealp{bolukbasi2016man}) allows us to ascribe any shifts in bias to the change in the frequency of one of the words. This simplifies the experiment and the conclusions we can draw from it, as compared to the case where multiple words are used in each context group.

With the embeddings trained on the resampled corpora and on the original corpus, we compute equation \ref{eq:BiasWE} on the words from the Glasgow Norms, a set of around 5,500 words with a score of gender association as perceived by human judgment \citep{scott2019glasgow}. See details in Appendix \ref{app:impact_bias}.

Our hypothesis is that bias can be heavily affected by the frequencies of the context words, so much so that answers to questions of the type "\emph{what are the most gender-biased words in this corpus?}" can be highly dependent on the frequencies of words in the corpus being studied.

\subsection{Results} \label{sec:impact_results}

The effect of frequencies on gender $\text{Bias}_{\operatorname{WE}}$ is different in each method (leftmost panel in Figure \ref{fig:impact_bias}). SGNS has an issue when measuring bias in high frequency words: in these words the association between $\text{Bias}_{\operatorname{WE}}$ and the frequency of $B$ is negative, while it is approximately constant for the rest of the frequency ranges. 


In FastText embeddings this negative association is observed across all frequency bins. Moreover, when the frequency of \textit{he} is low enough, all words have a positive (female) bias. This means the bias of specific words might appear to be high, when in fact the average bias of \textit{all} words is high.


Context words frequencies have the most influence on bias when using GloVe, as there are very different effects in each frequency bin. Both the level and ranking of bias are highly affected by frequency. More frequent target words tend to stick to the more frequent context word and less frequent words are attracted to the less frequent context. 

In summary, gender bias estimates can change substantially with the three methods even if the underlying distribution of co-occurrences remains constant. These shifts are triggered by the change in the context words' frequencies, which is an undesirable property in similarity measurements. In Figure \ref{fig:impact_gender_she} from Appendix \ref{app:impact_bias} we show that the frequency dependence persists when undersampling word $A$ instead of $B$. 

\subsubsection{Qualitative analysis of individual words} \label{sec:impact_words}

\begin{figure*}[!ht]
    \centering
    \includegraphics[width=\linewidth]{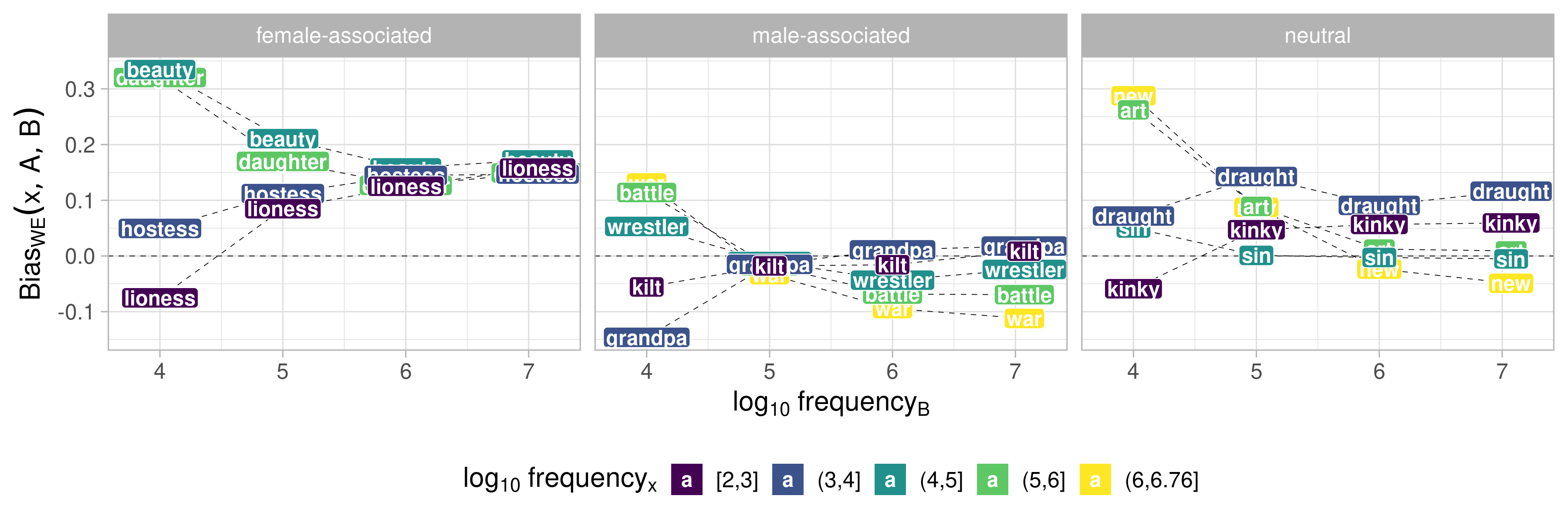}
    \caption{
        GloVe gender bias. Selected words are classified by frequency bin and by their perceived genderedness by human judgment ("female-associated", "male-associated", and "neutral"). Horizontal axes represent corpora where the frequency of $B$ (\textit{he}) varies and the frequency of $A$ (\textit{she}) is almost constant (see Table \ref{tab:undersampling_freqs_he}). \textbf{Biases can even change sign or ranking when frequencies change.}
        }
    \label{fig:impact_gender_words_glove}
\end{figure*}

To illustrate how this property can lead to misleading conclusions, we perform a qualitative analysis of the $\text{Bias}_{\operatorname{WE}}$ of individual words with GloVe, which seems to have the strongest frequency-based distortion.


We classify words by their perceived genderedness according to human judgment: "male" if their Glasgow gender norm is equal to or less than 2, "female" if it is 6 or higher, and "neutral" otherwise. For each class of words we sample a word for each one of five frequency bins and we study the changes in bias according to the frequency of \textit{he} in the corpus (Figure \ref{fig:impact_gender_words_glove}).

As observed in Figure \ref{fig:impact_bias}, the $\text{Bias}_{\operatorname{WE}}$ of frequent words (frequency $10^5$ onwards) is inversely correlated to the frequency of \textit{he}, while the association is positive for less frequent words (below $10^4$ occurrences). This occurs regardless of whether the words are perceived to be "male", "female" or "neutral". 

The bias is so dependent on the frequencies of the context words that the ranking of words can be reverted. For instance, if the frequency of \textit{he} is low enough, we might end up believing that "male-associated" words like \textit{war} and \textit{battle}, or "neutral" words as \textit{new} or \textit{art}, are female-biased words in our corpus, with even larger values than \textit{hostess}. At the same time, \textit{lioness} might tend to appear as male-biased, even more than \textit{wrestler}, \textit{battle} or \textit{war}. 

What is more, this behavior holds both for words that are inherently gendered, e.g. \textit{daughter} or \textit{grandpa}, and for words that are stereotypically associated with gender, e.g. \textit{beauty} or \textit{battle} (refer to Figure \ref{fig:impact_gender_words_glove_inherent_stereo} in Appendix \ref{app:impact_words} for a more detailed analysis). In Appendix \ref{app:impact_words} we also replicate the analysis of specific words for SGNS and FastText.

\subsubsection{Other biases}

To show that the findings do not apply to gender bias alone, we assess the impact on the measurement of bias along other cultural dimensions, namely ethnicity \citep{caliskan2017semantics, kozlowski2019geometry} and affluence \citep{kozlowski2019geometry}.

We measure ethnicity bias with $A=\{\textit{african}\}$ and $B=\{\textit{european}\}$, and affluence bias with $A=\{\textit{rich}\}$ and $B=\{\textit{poor}\}$. The words were chosen from \citet{kozlowski2019geometry} and aiming at reducing the ambiguity of the association being measured (e.g. african/european is less ambiguous than black/white).


We follow the same approach from section \ref{sec:impact_setup}. In these cases achieving the desired frequency levels implies also oversampling i.e. randomly replicating sentences containing $B$ (see Tables \ref{tab:resampling_freqs_race} and \ref{tab:resampling_freqs_affluence}). As with gender bias, we use the words from the Glasgow Norms as target words, applying the same filtering described in Appendix \ref{app:impact_bias}.

The findings are qualitatively the same as those in section \ref{sec:impact_results}: the estimates of bias can be highly dependent on the frequencies of the words involved, and the effect is different in each set of embeddings (middle and right panels of Figure \ref{fig:impact_bias}). Moreover, the affluence and ethnicity biases of each frequency range vary in a very similar manner to gender bias. This further supports the claim that the frequency of the words being compared heavily affects similarity scores and bias estimates, independently of the specific context words we have chosen to use in these experiments.


\begin{table}[H]
    \centering
    \begin{tabular}{|p{2.9cm}|C{1.7cm}|C{1.7cm}|}
        \hline
                                      & {\small \emph{african} ($A$)} & {\small \emph{european} ($B$)} \\ \hline
        {\small Oversampled Wiki.}    & $10^{5.39}$                   & $10^{6}$                       \\ \hline
        {\small Original Wiki.}       & $10^{5.37}$                   & $10^{5.58}$                    \\ \hline
        {\small Undersampled Wiki.} 1 & $10^{5.36}$                   & $10^{5}$                       \\ \hline
        {\small Undersampled Wiki.} 2 & $10^{5.36}$                   & $10^{4}$                       \\ \hline
    \end{tabular}
    \caption{
        Frequency of context words in the resampling experiment that either drops or replicates sentences with $B$ (\textit{european}). The frequency of word $A$ (\textit{african}) changes but to a minor extent.
        }
    \label{tab:resampling_freqs_race}
\end{table}

\begin{table}[H]
    \centering
    \begin{tabular}{|p{2.9cm}|C{1.7cm}|C{1.7cm}|}
        \hline
                                      & {\small \emph{rich} ($A$)} & {\small \emph{poor} ($B$)} \\ \hline
        {\small Oversampled Wiki.}    & $10^{5.05}$                & $10^{6}$                   \\ \hline
        {\small Original Wiki.}       & $10^{4.95}$                & $10^{5.21}$                \\ \hline
        {\small Undersampled Wiki. 1} & $10^{4.94}$                & $10^{5}$                   \\ \hline
        {\small Undersampled Wiki. 2} & $10^{4.93}$                & $10^{4}$                   \\ \hline
    \end{tabular}
    \caption{
        Frequency of context words in the resampling experiment that either drops or replicates sentences with $B$ (\textit{poor}). The frequency of word $A$ (\textit{rich}) changes but to a minor extent.
    }
    \label{tab:resampling_freqs_affluence}
\end{table}

\section{Discussion and conclusions} \label{sec:conclusion}

Static word embeddings are useful for computing semantic similarity between words since they capture the semantics of words. To assess biases in texts, computational social scientists frequently use word embedding similarity as a metric.

Static word embeddings can encode information on word frequency, according to earlier research. We examine the relationship between frequency and semantic similarity in SGNS, FastText, and GloVe embeddings in greater detail in this work. We find that the frequency of the words being compared affects their similarity score. This dependence is also present when words in the training corpus are randomly shuffled, demonstrating that the behavior is an artifact of the embeddings and not a result of actual associations found in the text. Moreover, we find that this frequency distortion persists under different hyperparameter settings.

In computational social science applications like bias measurement, the propensity of embeddings to encode frequency hampers their ability to measure semantic closeness. We conduct a controlled experiment that illustrates how measuring gender, affluence or ethnicity biases using embedding-based metrics might produce inaccurate results. The results indicate that the frequency of words can have a significant impact on the answers to questions like "\emph{what are the most gender biased terms?}" as biases can change sign or ranking when word frequencies are changed.

A way to mitigate the frequency distortion in embedding-based bias metrics could involve creating context groups $A$ and $B$ with words that have similar average frequencies, if possible. When doing this, frequency does not have a systematic effect in the sign of the subtraction of cosine similarities. Another approach is to randomly replicate documents prior to training embeddings so that the frequencies of $A$ and $B$ are balanced. However, it is worth noting that when measuring bias with context groups $A$ and $B$ with multiple words each, it can be challenging to achieve balanced frequencies for all words simultaneously.

\section*{Limitations} \label{sec:limitations}

Experiments were conducted solely on the English language. This means that our findings may not be directly applicable to languages that have more complex morphological features or richer grammatical genders. 

Even if our analyses are focused exclusively on the English Wikipedia corpus, we consider that the random-shuffling experiment is sufficiently generic to prove that the dependence on frequency would continue to hold true in other domains. 

The embedding-based metric we use to measure biases imply a binary understanding of stereotypes, which excludes other views. The context words were chosen from past studies and aiming at reducing the ambiguity of the associations being measured.


\bibliography{custom}

\begin{thebibliography}{28}
\expandafter\ifx\csname natexlab\endcsname\relax\def\natexlab#1{#1}\fi

\bibitem[{Aletras and Stevenson(2013)}]{aletras2013evaluating}
Nikolaos Aletras and Mark Stevenson. 2013.
\newblock \href {https://aclanthology.org/W13-0102} {Evaluating topic coherence using distributional semantics}.
\newblock In \emph{Proceedings of the 10th International Conference on Computational Semantics (IWCS 2013)--Long Papers}, pages 13--22.

\bibitem[{Altszyler et~al.(2017)Altszyler, Ribeiro, Sigman, and {Fernández Slezak}}]{altszyler2017interpretation}
Edgar Altszyler, Sidarta Ribeiro, Mariano Sigman, and Diego {Fernández Slezak}. 2017.
\newblock \href {https://doi.org/https://doi.org/10.1016/j.concog.2017.09.004} {The interpretation of dream meaning: Resolving ambiguity using latent semantic analysis in a small corpus of text}.
\newblock \emph{Consciousness and Cognition}, 56:178--187.

\bibitem[{Arora et~al.(2016)Arora, Li, Liang, Ma, and Risteski}]{arora2016latent}
Sanjeev Arora, Yuanzhi Li, Yingyu Liang, Tengyu Ma, and Andrej Risteski. 2016.
\newblock \href {https://doi.org/10.1162/tacl_a_00106} {A latent variable model approach to {PMI}-based word embeddings}.
\newblock \emph{Transactions of the Association for Computational Linguistics}, 4:385--399.

\bibitem[{Bojanowski et~al.(2017)Bojanowski, Grave, Joulin, and Mikolov}]{bojanowski2017enriching}
Piotr Bojanowski, Edouard Grave, Armand Joulin, and Tomas Mikolov. 2017.
\newblock \href {https://doi.org/10.1162/tacl_a_00051} {Enriching word vectors with subword information}.
\newblock \emph{Transactions of the Association for Computational Linguistics}, 5:135--146.

\bibitem[{Bolukbasi et~al.(2016)Bolukbasi, Chang, Zou, Saligrama, and Kalai}]{bolukbasi2016man}
Tolga Bolukbasi, Kai-Wei Chang, James~Y Zou, Venkatesh Saligrama, and Adam~T Kalai. 2016.
\newblock \href {https://proceedings.neurips.cc/paper/2016/file/a486cd07e4ac3d270571622f4f316ec5-Paper.pdf} {Man is to computer programmer as woman is to homemaker? debiasing word embeddings}.
\newblock In \emph{Advances in Neural Information Processing Systems}, volume~29. Curran Associates, Inc.

\bibitem[{Caliskan et~al.(2017)Caliskan, Bryson, and Narayanan}]{caliskan2017semantics}
Aylin Caliskan, Joanna~J. Bryson, and Arvind Narayanan. 2017.
\newblock \href {https://doi.org/10.1126/science.aal4230} {Semantics derived automatically from language corpora contain human-like biases}.
\newblock \emph{Science}, 356(6334):183--186.

\bibitem[{Charlesworth et~al.(2021)Charlesworth, Yang, Mann, Kurdi, and Banaji}]{charlesworth2021gender}
Tessa~ES Charlesworth, Victor Yang, Thomas~C Mann, Benedek Kurdi, and Mahzarin~R Banaji. 2021.
\newblock \href {https://doi.org/10.1177/0956797620963619} {Gender stereotypes in natural language: Word embeddings show robust consistency across child and adult language corpora of more than 65 million words}.
\newblock \emph{Psychological Science}, 32(2):218--240.

\bibitem[{DeFranza et~al.(2020)DeFranza, Mishra, and Mishra}]{defranza2020language}
David DeFranza, Himanshu Mishra, and Arul Mishra. 2020.
\newblock \href {https://doi.org/10.1037/pspa0000188} {How language shapes prejudice against women: An examination across 45 world languages.}
\newblock \emph{Journal of Personality and Social Psychology}, 119(1):7--22.

\bibitem[{Diuk et~al.(2012)Diuk, Slezak, Raskovsky, Sigman, and Cecchi}]{diuk2012quantitative}
Carlos~G Diuk, D~Fernandez Slezak, Iv{\'a}n Raskovsky, Mariano Sigman, and Guillermo~A Cecchi. 2012.
\newblock \href {https://doi.org/10.3389/fnint.2012.00080} {A quantitative philology of introspection}.
\newblock \emph{Frontiers in integrative neuroscience}, 6:80.

\bibitem[{Ethayarajh(2019)}]{ethayarajh2019contextual}
Kawin Ethayarajh. 2019.
\newblock \href {https://doi.org/10.18653/v1/D19-1006} {How contextual are contextualized word representations? {C}omparing the geometry of {BERT}, {ELM}o, and {GPT}-2 embeddings}.
\newblock In \emph{Proceedings of the 2019 Conference on Empirical Methods in Natural Language Processing and the 9th International Joint Conference on Natural Language Processing (EMNLP-IJCNLP)}, pages 55--65, Hong Kong, China. Association for Computational Linguistics.

\bibitem[{Garg et~al.(2018)Garg, Schiebinger, Jurafsky, and Zou}]{garg2018word}
Nikhil Garg, Londa Schiebinger, Dan Jurafsky, and James Zou. 2018.
\newblock \href {https://doi.org/10.1073/pnas.1720347115} {Word embeddings quantify 100 years of gender and ethnic stereotypes}.
\newblock \emph{Proceedings of the National Academy of Sciences}, 115(16):E3635--E3644.

\bibitem[{Gong et~al.(2018)Gong, He, Tan, Qin, Wang, and Liu}]{gong2018frage}
Chengyue Gong, Di~He, Xu~Tan, Tao Qin, Liwei Wang, and Tie-Yan Liu. 2018.
\newblock \href {https://proceedings.neurips.cc/paper/2018/file/e555ebe0ce426f7f9b2bef0706315e0c-Paper.pdf} {Frage: Frequency-agnostic word representation}.
\newblock In \emph{Advances in Neural Information Processing Systems}, volume~31. Curran Associates, Inc.

\bibitem[{Hellrich and Hahn(2016)}]{hellrich2016bad}
Johannes Hellrich and Udo Hahn. 2016.
\newblock \href {https://aclanthology.org/C16-1262} {Bad {C}ompany{---}{N}eighborhoods in neural embedding spaces considered harmful}.
\newblock In \emph{Proceedings of {COLING} 2016, the 26th International Conference on Computational Linguistics: Technical Papers}, pages 2785--2796, Osaka, Japan. The COLING 2016 Organizing Committee.

\bibitem[{Jones et~al.(2020)Jones, Amin, Kim, and Skiena}]{jones2020stereotypical}
Jason~J Jones, Mohammad~Ruhul Amin, Jessica Kim, and Steven Skiena. 2020.
\newblock \href {https://doi.org/10.15195/v7.a1} {Stereotypical gender associations in language have decreased over time}.
\newblock \emph{Sociological Science}, 7:1--35.

\bibitem[{Kozlowski et~al.(2019)Kozlowski, Taddy, and Evans}]{kozlowski2019geometry}
Austin~C. Kozlowski, Matt Taddy, and James~A. Evans. 2019.
\newblock \href {https://doi.org/10.1177/0003122419877135} {The geometry of culture: Analyzing the meanings of class through word embeddings}.
\newblock \emph{American Sociological Review}, 84(5):905--949.

\bibitem[{Levy et~al.(2015)Levy, Goldberg, and Dagan}]{levy2015improving}
Omer Levy, Yoav Goldberg, and Ido Dagan. 2015.
\newblock \href {https://doi.org/10.1162/tacl_a_00134} {Improving distributional similarity with lessons learned from word embeddings}.
\newblock \emph{Transactions of the Association for Computational Linguistics}, 3:211--225.

\bibitem[{Lewis and Lupyan(2020)}]{lewis2020gender}
Molly Lewis and Gary Lupyan. 2020.
\newblock \href {https://doi.org/10.1038/s41562-020-0918-6} {Gender stereotypes are reflected in the distributional structure of 25 languages}.
\newblock \emph{Nature Human Behaviour}, 4(10):1021--1028.

\bibitem[{Mikolov et~al.(2013)Mikolov, Sutskever, Chen, Corrado, and Dean}]{mikolov2013distributed}
Tomas Mikolov, Ilya Sutskever, Kai Chen, Greg~S Corrado, and Jeff Dean. 2013.
\newblock \href {https://proceedings.neurips.cc/paper/2013/file/9aa42b31882ec039965f3c4923ce901b-Paper.pdf} {Distributed representations of words and phrases and their compositionality}.
\newblock In \emph{Advances in Neural Information Processing Systems}, volume~26. Curran Associates, Inc.

\bibitem[{Mimno and Thompson(2017)}]{mimno2017strange}
David Mimno and Laure Thompson. 2017.
\newblock \href {https://doi.org/10.18653/v1/D17-1308} {The strange geometry of skip-gram with negative sampling}.
\newblock In \emph{Proceedings of the 2017 Conference on Empirical Methods in Natural Language Processing}, pages 2873--2878, Copenhagen, Denmark. Association for Computational Linguistics.

\bibitem[{Mota et~al.(2022)Mota, Soares, Altszyler, S{\'a}nchez-Gendriz, Muto, Heib, Slezak, Sigman, Copelli, Schabus et~al.}]{mota2022imagetic}
Nat{\'a}lia~Bezerra Mota, Ernesto Soares, Edgar Altszyler, Ignacio S{\'a}nchez-Gendriz, Vincenzo Muto, Dominik Heib, Diego~F Slezak, Mariano Sigman, Mauro Copelli, Manuel Schabus, et~al. 2022.
\newblock \href {https://doi.org/https://doi.org/10.1016/j.neuroimage.2022.119690} {Imagetic and affective measures of memory reverberation diverge at sleep onset in association with theta rhythm}.
\newblock \emph{NeuroImage}, page 119690.

\bibitem[{Mu and Viswanath(2018)}]{mu2018allbutthetop}
Jiaqi Mu and Pramod Viswanath. 2018.
\newblock \href {https://openreview.net/forum?id=HkuGJ3kCb} {All-but-the-top: Simple and effective postprocessing for word representations}.
\newblock In \emph{International Conference on Learning Representations}.

\bibitem[{Pennington et~al.(2014)Pennington, Socher, and Manning}]{pennington2014glove}
Jeffrey Pennington, Richard Socher, and Christopher~D. Manning. 2014.
\newblock \href {http://www.aclweb.org/anthology/D14-1162} {Glove: Global vectors for word representation}.
\newblock In \emph{Empirical Methods in Natural Language Processing (EMNLP)}, pages 1532--1543.

\bibitem[{{\v R}eh{\r u}{\v r}ek and Sojka(2010)}]{rehurek2010gensim}
Radim {\v R}eh{\r u}{\v r}ek and Petr Sojka. 2010.
\newblock {Software Framework for Topic Modelling with Large Corpora}.
\newblock In \emph{{Proceedings of the LREC 2010 Workshop on New Challenges for NLP Frameworks}}, pages 45--50, Valletta, Malta. ELRA.
\newblock \url{http://is.muni.cz/publication/884893/en}.

\bibitem[{Schnabel et~al.(2015)Schnabel, Labutov, Mimno, and Joachims}]{schnabel2015evaluation}
Tobias Schnabel, Igor Labutov, David Mimno, and Thorsten Joachims. 2015.
\newblock \href {https://doi.org/10.18653/v1/D15-1036} {Evaluation methods for unsupervised word embeddings}.
\newblock In \emph{Proceedings of the 2015 Conference on Empirical Methods in Natural Language Processing}, pages 298--307, Lisbon, Portugal. Association for Computational Linguistics.

\bibitem[{Scott et~al.(2019)Scott, Keitel, Becirspahic, Yao, and Sereno}]{scott2019glasgow}
{Graham G} Scott, Anne Keitel, Marc Becirspahic, Bo~Yao, and {Sara C} Sereno. 2019.
\newblock \href {https://doi.org/10.3758/s13428-018-1099-3} {The {G}lasgow {N}orms: Ratings of 5,500 words on nine scales}.
\newblock \emph{Behavior Research Methods}, 51:1258--1270.

\bibitem[{Sezerer and Tekir(2021)}]{sezerer2021survey}
Erhan Sezerer and Selma Tekir. 2021.
\newblock \href {https://doi.org/10.48550/ARXIV.2110.01804} {A survey on neural word embeddings}.
\newblock \emph{arXiv preprint}, arXiv:2110.01804.

\bibitem[{Valentini et~al.(2022)Valentini, Rosati, Slezak, and Altszyler}]{valentini2022undesirable}
Francisco Valentini, Germ{\'a}n Rosati, Diego~Fernandez Slezak, and Edgar Altszyler. 2022.
\newblock The undesirable dependence on frequency of gender bias metrics based on word embeddings.
\newblock In \emph{Findings of the Association for Computational Linguistics: EMNLP 2022}. Association for Computational Linguistics.

\bibitem[{Wilson and Schakel(2015)}]{wilson2015controlled}
Benjamin~J Wilson and Adriaan~MJ Schakel. 2015.
\newblock \href {https://doi.org/10.48550/ARXIV.1510.02675} {Controlled experiments for word embeddings}.
\newblock \emph{arXiv preprint arXiv:1510.02675}.

\end{thebibliography}
\bibliographystyle{acl_natbib}

\appendix

\section{Data and methods} \label{app:methods}

We build the Wikipedia corpus from the April 2021 English Wikipedia dump (\url{https://archive.org/download/enwiki-20210401}, license CC BY-SA 3.0). Wikipedia is freely available, easily accessible and has been used in previous experiments \citep{levy2015improving}. We remove articles with less than 50 words. Pre-processing includes sentence splitting, lowercasing and removing non alpha-numeric symbols, and produces a corpus of 78 million sentences and 1.2 billion tokens.

We train word embeddings with 300 dimensions. All words with less than 100 occurrences are removed before obtaining word-context pairs and we use a sliding window size of 10 tokens by default. SGNS and FastText are trained with Gensim's implementation (\citealp{rehurek2010gensim}, v4.2.0, licensed under GNU LGPLv2.1), and GloVe is trained with \citet{pennington2014glove}'s implementation (v1.2, Apache License Version 2.0). 

We used a desktop computer with 16 cores Intel Core i7-11700 CPU and 32GB RAM. Depending on the corpus being used, training took between 0.5 and 1.2 hours per epoch with SGNS and FastText, and 5 and 15 minutes per iteration with GloVe.

\section{Euclidean distance-based similarity metric} \label{app:euclidean}

\begin{figure*}
    \centering
    \includegraphics[width=\linewidth]{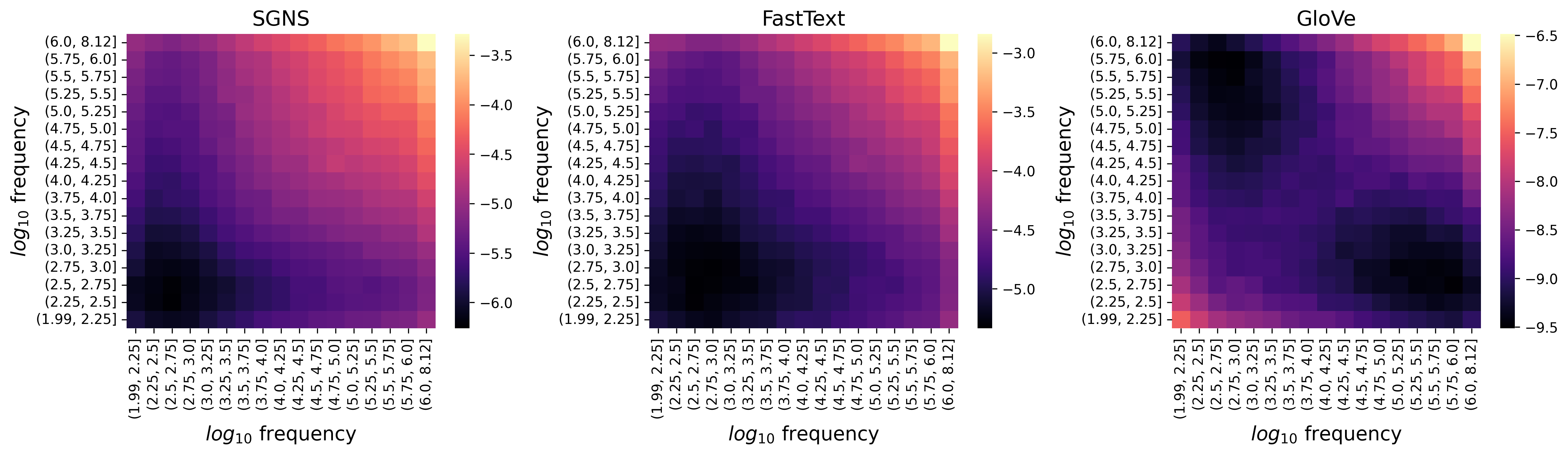}
    \caption{
        Mean negative Euclidean distance between 500 random word pairs for each combination of frequencies in embeddings trained on Wikipedia.
    }
    \label{fig:euclidian_unshuffled}
\end{figure*}

\begin{figure*}
    \centering
    \includegraphics[width=\linewidth]{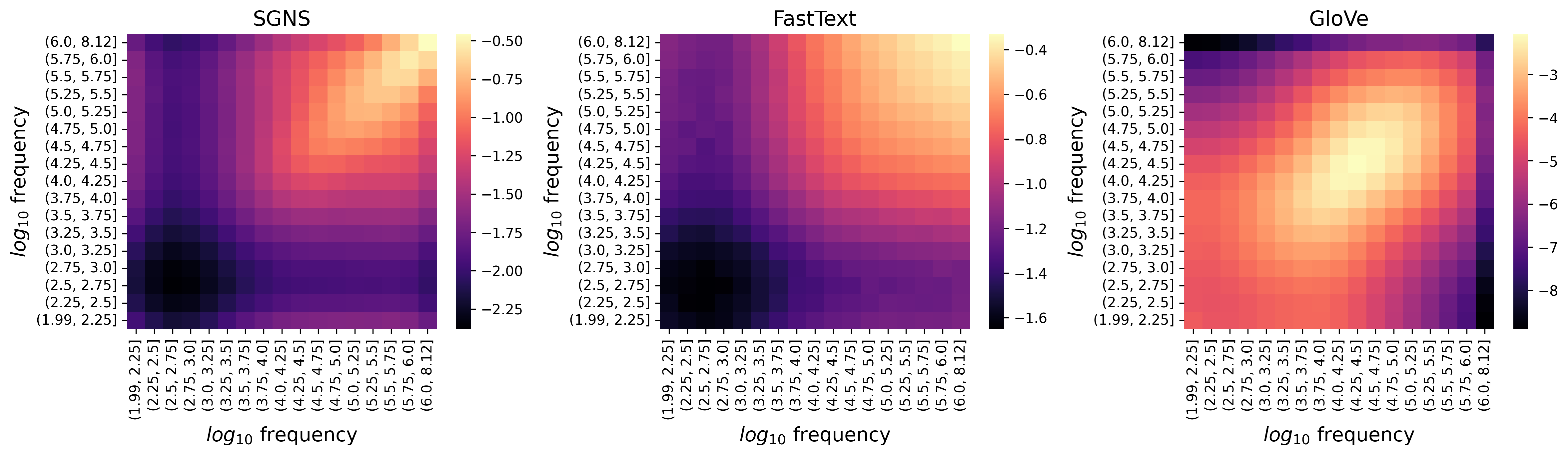}
    \caption{
        Mean negative Euclidean distance between 500 random word pairs for each combination of frequencies in embeddings trained on a shuffled version of Wikipedia.
    }
    \label{fig:euclidian_shuffled}
\end{figure*}

Figures \ref{fig:euclidian_unshuffled} and \ref{fig:euclidian_shuffled} show the association between similarity and word frequency when using a similarity measure based on Euclidean distance. 

\begin{figure*}
    \centering
    \includegraphics[width=0.5\linewidth]{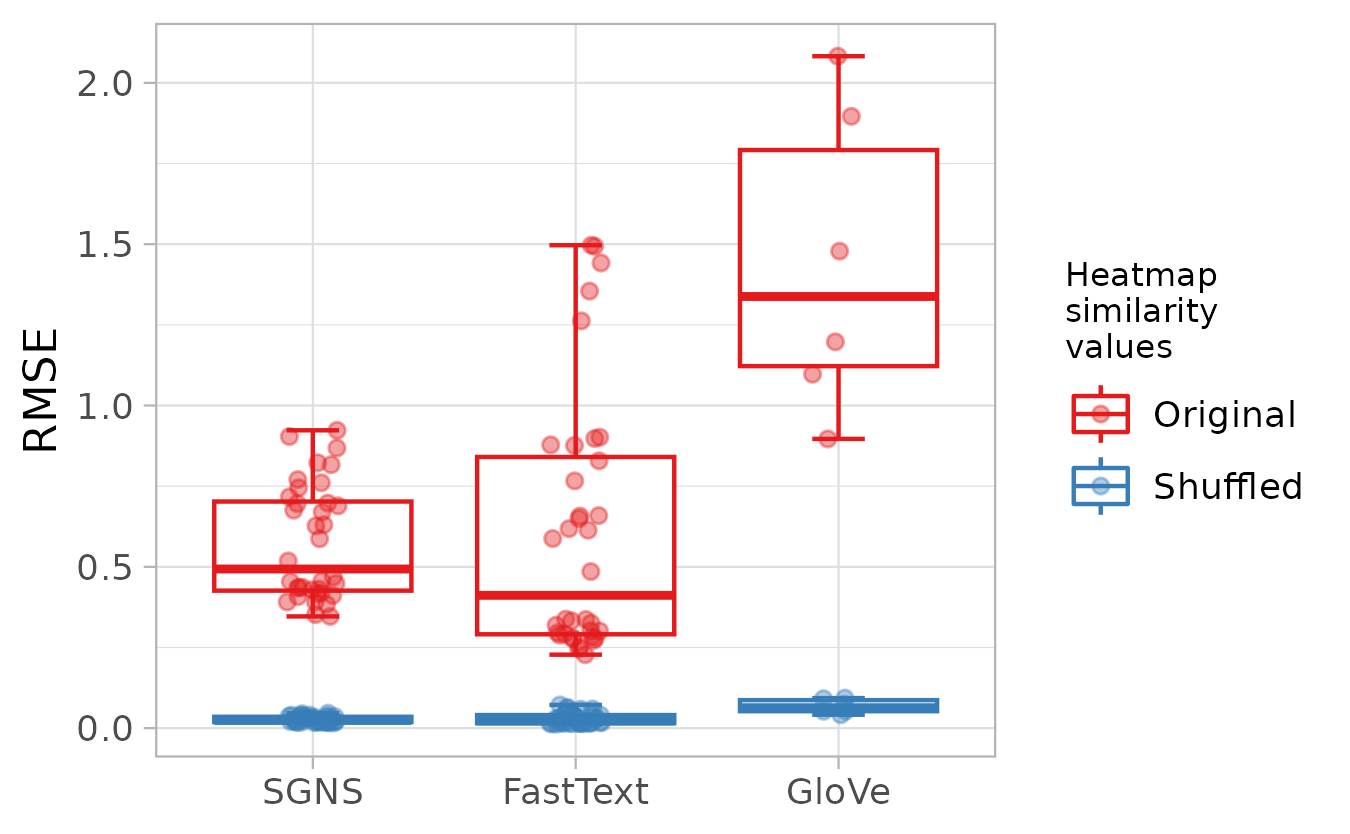}
    \caption{
    Association between frequency and negative Euclidean distance as measured by the RMSE metric. Each point is a hyperparameter setting of embeddings trained on shuffled Wikipedia. The RMSE computed with the actual heatmaps (red) is compared to the RMSE computed with the heatmaps obtained by shuffling the similarity values of the pairs of words (blue).
    }
    \label{fig:hyperparams_euclidean}
\end{figure*}

Figure \ref{fig:hyperparams_euclidean} shows the distribution of the RMSE metric across different hyperparameter settings when using the Euclidean-based measure instead of cosine similarity. The relatively high values of RMSE in red as compared to the baseline values in blue shows that the frequency-based distortion is present in all hyperparameter settings even when using Euclidean distance.

\section{Principal Components Analysis} \label{app:pca}

\begin{figure*}
    \centering
    \includegraphics[width=\linewidth]{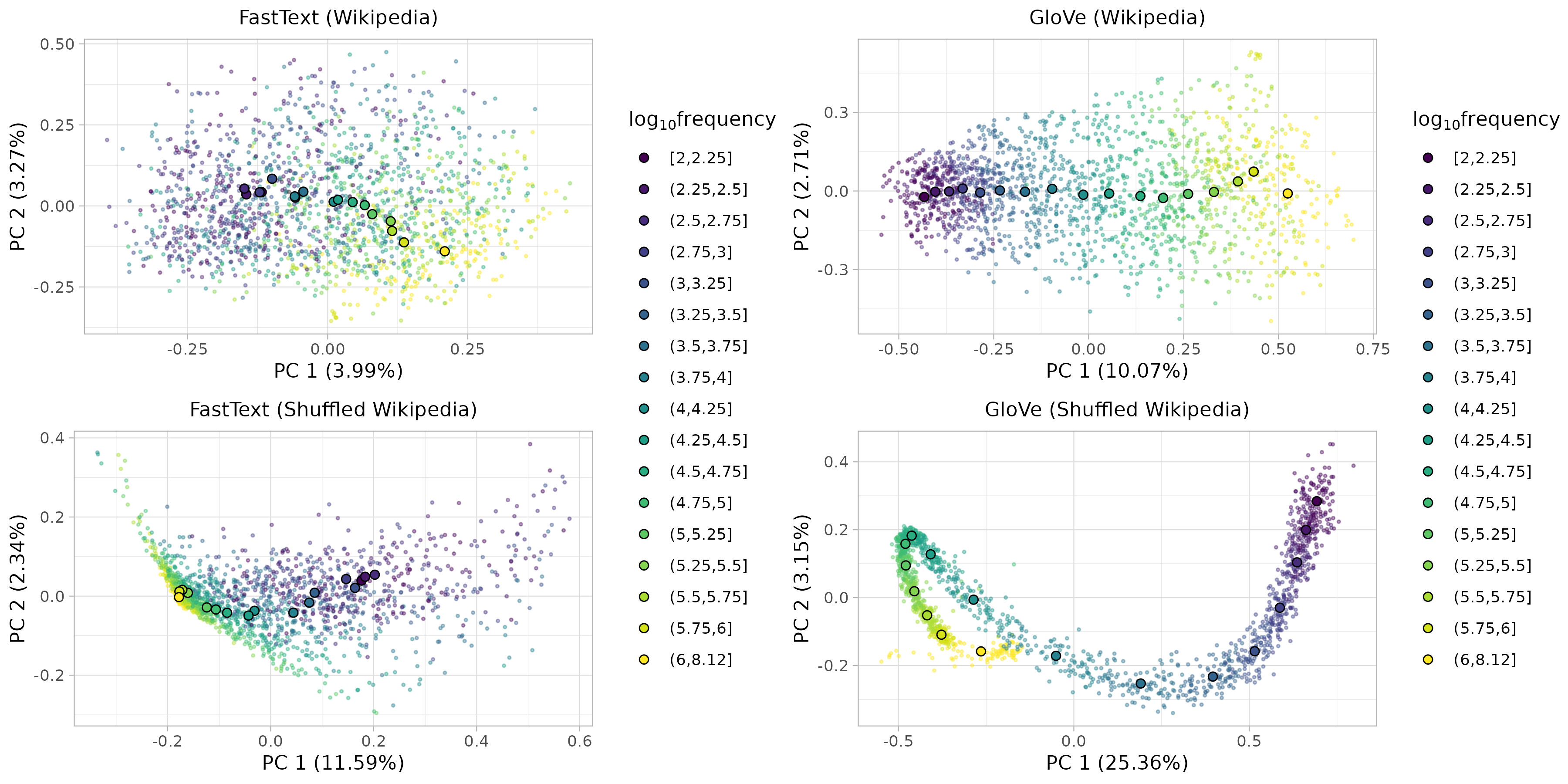}
    \caption{
        Top principal components of a sample of embeddings stratified by frequency trained on the original Wikipedia (top) and the shuffled Wikipedia (bottom), for FastText (left) and GloVe (right). Vectors are normalized to unit length before PCA and there are 100 words by frequency bin. Centroids are displayed with larger markers.
        }
    \label{fig:pca_centroids}
\end{figure*}

Figure \ref{fig:pca_centroids} shows the top two principal components of each frequency bin in unshuffled Wikipedia and shuffled Wikipedia for embeddings trained with FastText and GloVe.

\section{Sensitivity to hyperparameters} \label{app:hyperparameters}

\begin{table}[H]
    \centering
    \begin{tabularx}{\linewidth}{X X X X X}
        \hline
                         & \multicolumn{2}{c}{SGNS+FT+GloVe} & \multicolumn{2}{c}{SGNS+FT}                          \\
        \cline{2-3} \cline{4-5}
                         & Cos.                              & Eucl.                       & Cos.    & Eucl.        \\
        \hline
        \texttt{win=}5   & $0.01$                            & $0.03$                      & $0.01$  & $0.04$       \\
        \texttt{win=}10  & $0.02$                            & $0.10$                      & $0.02$  & $0.13$       \\
        \texttt{w+c=}yes & $0.01$                            & $0.26^{***}$                & $0.01$  & $0.24^{***}$ \\
        \texttt{neg=}5   &                                   &                             & $-0.01$ & $0.03$       \\
        \texttt{neg=}15  &                                   &                             & $-0.03$ & $0.04$       \\
        \texttt{cds=}1   &                                   &                             & $-0.00$ & $-0.04$      \\
        \hline
        N                & $156$                             & $156$                       & $144$   & $144$        \\
        \hline
        \multicolumn{5}{l}{\scriptsize{$^{***}p<0.001$; $^{**}p<0.01$; $^{*}p<0.05$}}
    \end{tabularx}
    \caption{Coefficients of the linear regressions between the RMSE metric and hyperparameter choices, measuring embedding similarity with both cosine ("Cos.") and negative euclidean distance ("Eucl."). \textbf{No hyperparameter choice significantly affects the association between frequency and cosine similarity.}}
    \label{tab:regression_hps}
\end{table}

Table \ref{tab:regression_hps} displays the coefficients of a linear model with the hyperparameter values as predictors and the RMSE as the outcome variable, considering the cosine and the Euclidean-based similarity metrics. The only hyperparameter choice that tends to systematically yield higher RMSE values (stronger association between frequency and similarity) is adding context vectors (\texttt{w+c} = \emph{yes}), but only when using negative Euclidean distance as similarity metric.

\section{Impact on bias measurement} \label{app:impact_bias}

The Glasgow Norms \citep{scott2019glasgow} comprise a set of 5,553 English words rated by subjects who were asked to measure the degree to which each word is associated with male or female behavior on a scale from 1 (feminine) to 7 (masculine). We flip the scale so that the norm represents femaleness according to human judgment. 

We discard the norms of homonyms and of words with uppercase characters. Moreover, we only consider words that are in the vocabularies of all embeddings trained on the original corpus and the resampled corpora. Finally, we drop any words that change their frequency bin between corpora. This results in a set of 4,384, 4,641, and 4,617 words to measure gender, ethnicity and affluence bias in Figure \ref{fig:impact_bias}.

\begin{figure*}
    \centering
    \includegraphics[width=\linewidth]{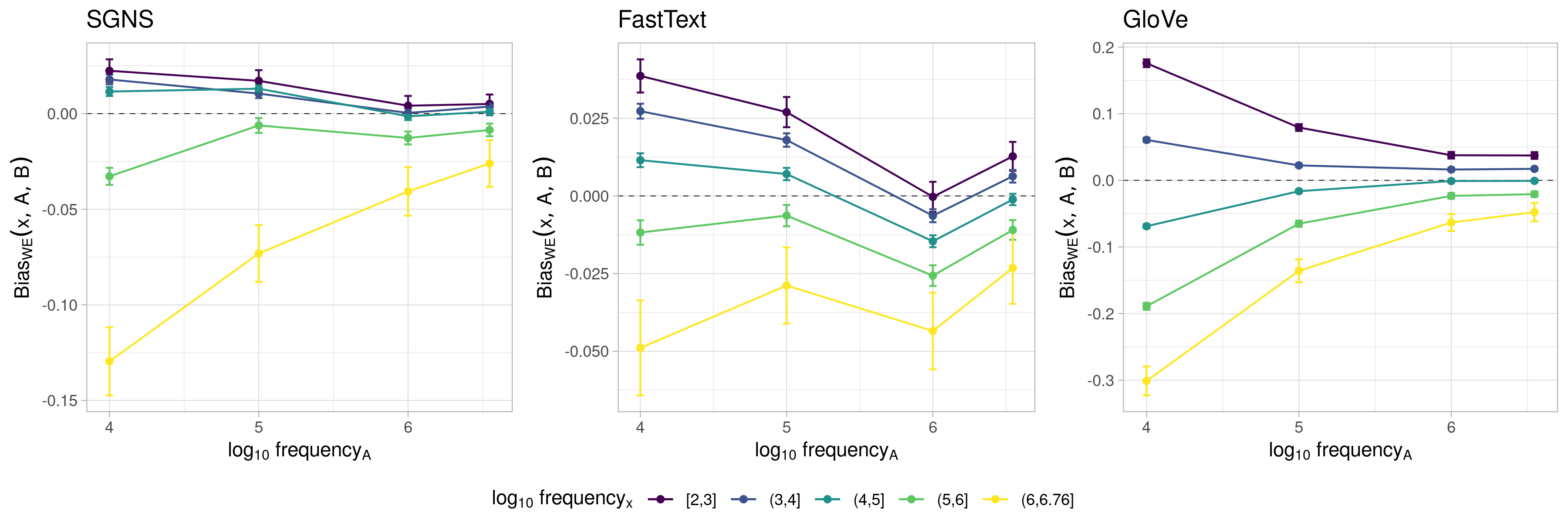}
    \caption{
        Female/male gender bias of 4,384 target words grouped by frequency. Horizontal axes represent corpora where the frequency of $A$ (\textit{she}) varies and the frequency of $B$ (\textit{he}) is almost constant (see Table \ref{tab:undersampling_freqs_she}). Mean bias is plotted with bootstrap confidence intervals.
        }
    \label{fig:impact_gender_she}
\end{figure*}

Figure \ref{fig:impact_gender_she} shows the effect of undersampling word $A$ (\textit{she}) instead of $B$ (\textit{he}). The frequencies employed in this experiment are in Table \ref{tab:undersampling_freqs_she}. In the same manner as in section \ref{sec:impact_results}, GloVe exhibits the highest frequency-based distortion, as more frequent target words stick to the more frequent context word (here, \textit{he}) and less frequent words are attracted to the less frequent context (\textit{she}). SGNS also presents the same effect in high frequency words as the one observed in Figure \ref{fig:impact_bias}. The main difference with respect to the experiment in section \ref{sec:impact_results} occurs with FastText. We have no hypothesis about the reason for this discrepancy.

    \begin{table}[H]
        \centering
        \begin{tabular}{|l|c|c|}
            \hline
                                & \emph{she} ($A$) & \emph{he} ($B$) \\ \hline
            Original Wiki.       & $10^{6.55}$      & $10^{7.07}$     \\ \hline
            Undersampled Wiki. 1 & $10^{6}$         & $10^{7.07}$     \\ \hline
            Undersampled Wiki. 2 & $10^{5}$         & $10^{7.07}$     \\ \hline
            Undersampled Wiki. 3 & $10^{4}$         & $10^{7.07}$     \\ \hline
        \end{tabular}
        \caption{
        Frequency of context words in the undersampling experiment that drops sentences with word $A$ (\emph{she}). The frequency of $B$ (\emph{he}) decreases but to a minor extent.
        }
        \label{tab:undersampling_freqs_she}
    \end{table}

\subsection{Qualitative analysis of individual words (SGNS and FastText)} \label{app:impact_words}

\begin{figure*}
    \centering
    \includegraphics[width=\linewidth]{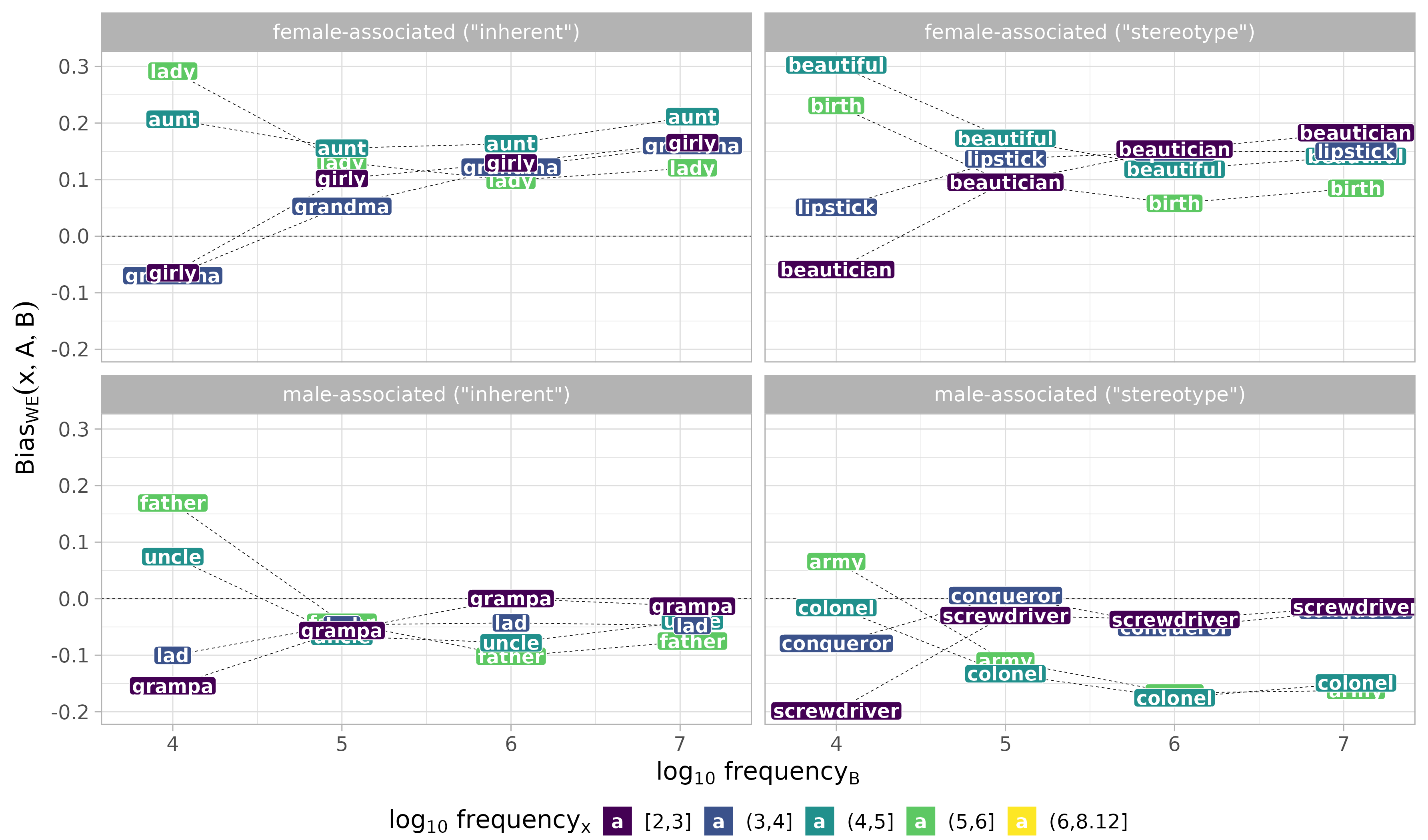}
    \caption{GloVe female/male gender bias of inherently and stereotypically gendered words in the experiment that undersamples $B$ (\textit{he}).}
    \label{fig:impact_gender_words_glove_inherent_stereo}
\end{figure*}

We chose male and female-associated words that are either inherently gendered or stereotypically associated with gender and studied the behavior of bias in the experiment that undersamples \textit{he} (Figure \ref{fig:impact_gender_words_glove_inherent_stereo}). Results reveal that the frequency distortion affects the bias of both types of words in the same way.

\begin{figure*}
    \centering
    \includegraphics[width=\linewidth]{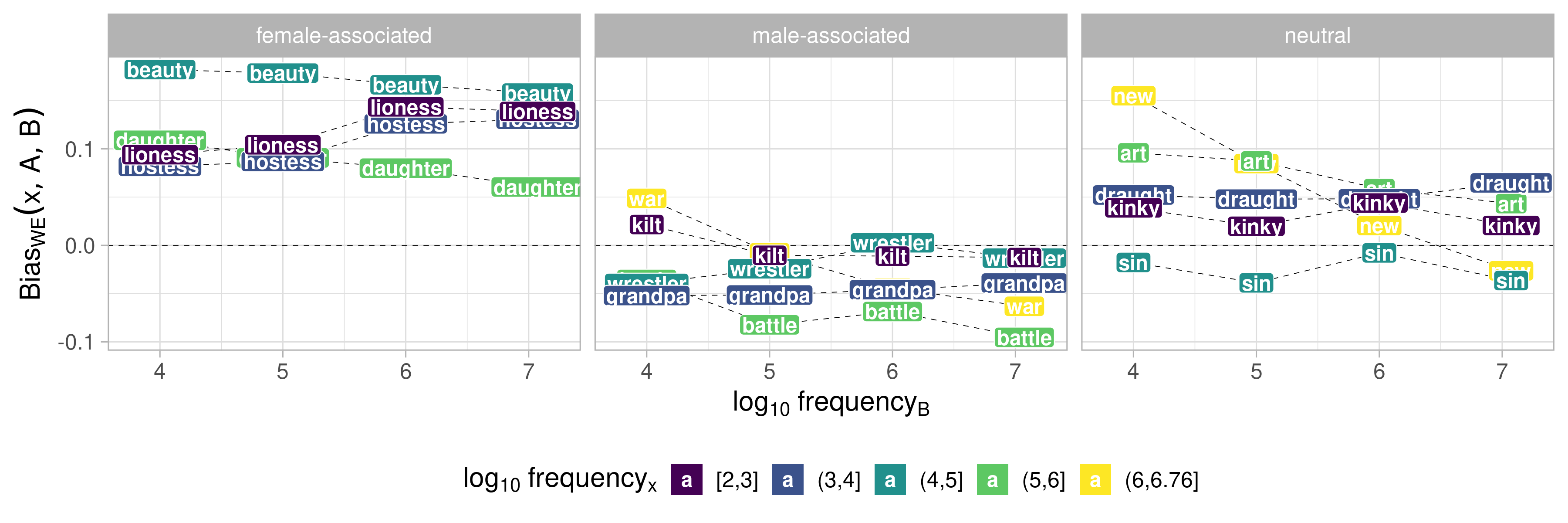}
    \caption{SGNS female/male gender bias of words in the experiment that undersamples $B$ (\textit{he}).}
    \label{fig:impact_gender_words_sgns}
\end{figure*}

\begin{figure*}
    \centering
    \includegraphics[width=\linewidth]{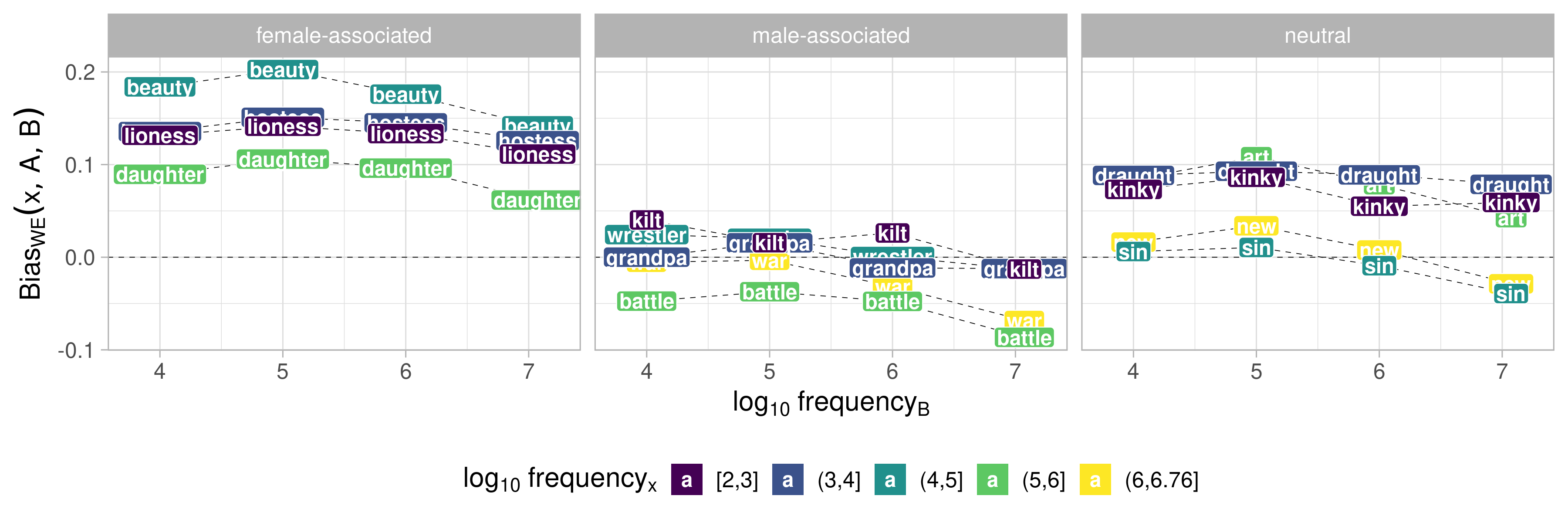}
    \caption{FastText female/male gender bias of words in the experiment that undersamples $B$ (\textit{he}).}
    \label{fig:impact_gender_words_fasttext}
\end{figure*}

In Figures \ref{fig:impact_gender_words_sgns} and \ref{fig:impact_gender_words_fasttext} we replicate the analysis of specific words for SGNS and FastText, respectively. These words are a subset of the words used to make the leftmost panel of Figure \ref{fig:impact_bias}.

\end{document}